\documentclass[sigconf,screen]{acmart}
\AtBeginDocument{%
  \providecommand\BibTeX{{%
    \normalfont B\kern-0.5em{\scshape i\kern-0.25em b}\kern-0.8em\TeX}}}

\setcopyright{rightsretained}

\makeatletter
\@acmownedfalse
\makeatother

\acmConference[MuC'22]{Veröffentlicht durch die Gesellschaft für Informatik e.V.\\
in K. Marky, U. Grünefeld \& T. Kosch (Hrsg.):\\ 
Mensch und Computer 2022 – Workshopband}{04.-07. September 2022}{Darmstadt}
\acmISBN{}
\acmPrice{}

\acmDOI{10.18420/muc2022-mci-ws13-453}

\usepackage[nolist]{acronym}
\usepackage{subcaption}
\usepackage{balance}
\usepackage{flushend}
\begin{document}

\title[Exploring Human-robot Interaction by Simulating Robots]{Exploring Human-robot Interaction by Simulating Robots}

\author{Khaled Kassem}
\affiliation{%
  \institution{TU Wien}
  \streetaddress{Favoritenstr. 9-11}
  \city{Vienna}
  \country{Austria}}
\email{khaled.k.kassem@tuwien.ac.at}
\author{Florian Michahelles}
\affiliation{%
  \institution{TU Wien}
  \streetaddress{Favoritenstr. 9-11}
  \city{Vienna}
  \country{Austria}}
\email{florian.michahelles@tuwien.ac.at}

\renewcommand{\shortauthors}{Kassem et al.}

\begin{abstract}
 As collaborative robots enter industrial shop floors, logistics, and manufacturing, rapid and flexible evaluation of human-machine interaction has become more important. The availability of consumer headsets for virtual and augmented realities has lowered the barrier of entry for virtual environments. In this paper, we explore the different aspects of using such environments for simulating robots in user studies and present the first findings from our own research work. Finally, we recommend directions for applying and using simulation in human-robot interaction.
\end{abstract}

\begin{acronym}
\acro{hci}[HCI]{Human-Computer Interaction}
\acro{hri}[HRI]{Human-Robot Interaction}
\acro{hrc}[HRC]{Human-Robot Collaboration}
\acro{ar}[AR]{Augmented Reality}
\acro{vr}[VR]{Virtual Reality}
\acro{xr}[XR]{Mixed Reality}
\acro{ems}[EMS]{Electrical Muscle Stimulation}
\acro{ros}[ROS]{Robot Operating System}
\acro{ct} [CT] {Collaborative Task}
\acro{eeg} [EEG] {electroencephalogram}
\end{acronym}

\newcommand{\red}[1]{\textcolor{red}{#1}}

\maketitle

\section{Introduction}
Developments in sensor technology and manufacturing processes have led to the creation of small and safe robots (referred to as cobots) that can safely physically interact with a human operator in a shared workspace~\cite{akella}. Cobots are a new and growing market, with value expected to reach \$9.1 Billion by 2025~\cite{munster_2019}. This growth of demand on robots leads to \ac{hri} becoming an important field of scientific inquiry~\cite{song2017}. A user-centred approach to \ac{hri} research is important for building safe and usable robots~\cite{heinzmann1999safe}. HRI user research relies on precise recording and analysis of human interactions with robots. This interaction can be with real or  simulated robots in controlled user studies~\cite{groechel22}. Research relying on virtual settings for HRI has been supported by the recent boom in the use of mixed-reality consumer headsets~\cite{groechel22}. In this paper we focus of the research question \textbf{"What are the benefits of evaluating human-robot interaction with simulated robots?"}, and we reflect on this question with findings from our work.
\section{Approach}
Simulating a robot for a study relies on three components, namely: virtualization, control, and the supporting hardware.

\textbf{Virtualization}
Tools like Unity3D~\footnote{https://unity.com/} have made building a virtual environment much easier and more accessible, especially with the rise of game development. Such tools allow flexibility to the researcher to experiment with different types of robots without needing to have access to all the different physical models, such as humanoid robots or collaborative robotic arms of varying degrees of freedom. A simulated robot also enables running studies in different environments, which could be real (with augmented reality) or can themselves be simulated as well (such as in a \ac{vr} setting). Additionally, Unified Robotics Description Format (URDF) files are widely available for a wide variety or robot classes. URDF files can then be used to generate 3D models of robots which can in turn be made into templates, or \textit{prefabricated game objects} ("prefabs" for short). Other prefabs of other objects can be used to rapidly setup environment elements such as furniture, panels, surfaces, and walls.

\textbf{Control}
Controlling a robot involves path and trajectory planning as well as forward and inverse kinematics for the precise control of joint positions and angles. A simulated robot that behaves similarly to a real robot requires the same form of control for proper behavior. Tools like the Robot Operating System~\footnote{https://www.ros.org/} (ROS) enable the control of a robot, provided a valid URDF model. Special-purpose plugins exist for establishing a connection between Unity3D and ROS in order to control a simulated robot.

\textbf{Hardware}
Commercial mixed reality headsets such as the Microsoft Hololens~\footnote{https://www.microsoft.com/en-us/hololens/hardware} for \ac{ar}, and the Oculus Quest~\footnote{https://store.facebook.com/at/en/quest/products/quest-2/} for \ac{vr} can be used to interact with a simulated robot in a real or virtual environment. The Oculus Quest headsets, for example, are equipped with hand-held controllers, haptic feedback, stereo audio, as well as hand recognition. These tools allow multiple interactions modality between the user and the simulation.
\section{Case Studies}
In the following part we provide specific examples of virtual environments we developed for our studies. We use the case studies as demonstration for what we could create for user studies in simulation. All environments were developed with Unity3D and linked to ROS. Figures (1-3) show screenshots taken from different environments built for the studies as they appeared in Unity3D. We discuss the findings derived from our studies in section \ref{sec:discussion}. 

\begin{figure}[h]
    \includegraphics[width=0.9\linewidth]{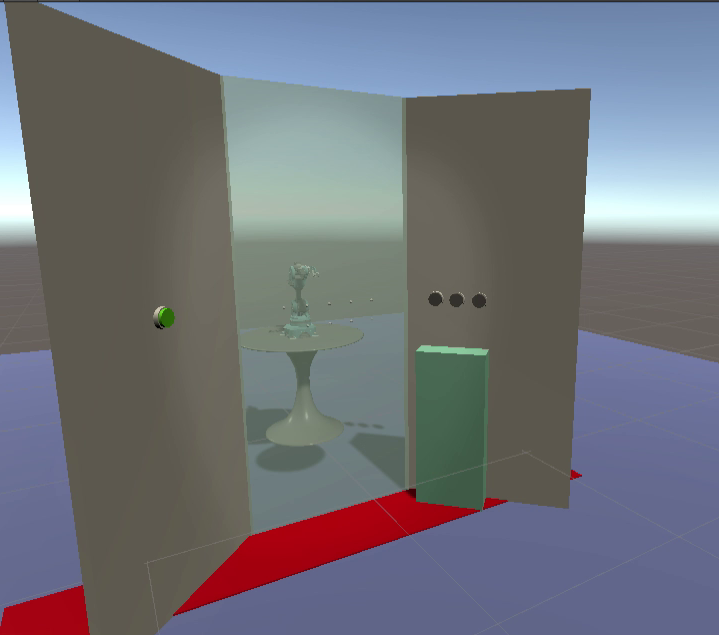}
    \caption{Behind the Wall: The user cooperates with a cobot sitting behind a wall. The wall becomes opaque and the user loses line of sight with the robot}
    \label{fig:behindthewall}
\end{figure}

\begin{figure}[h]
    \includegraphics[width=0.9\linewidth]{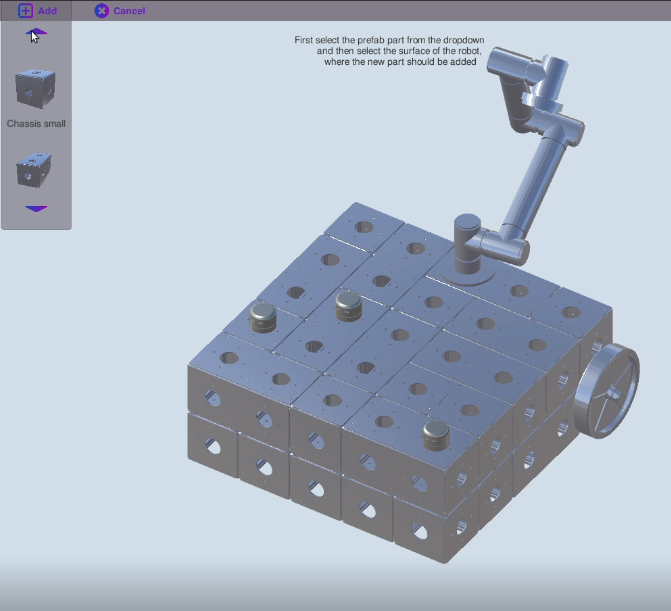}
    \caption{Modular Robot. The lower platform consists of multiple cubic modules. A drop down menu to add the individual modules can be seen on the top left.}
    \label{fig:modularRobot}
\end{figure}

\begin{figure}[h]
    \includegraphics[width=0.9\linewidth]{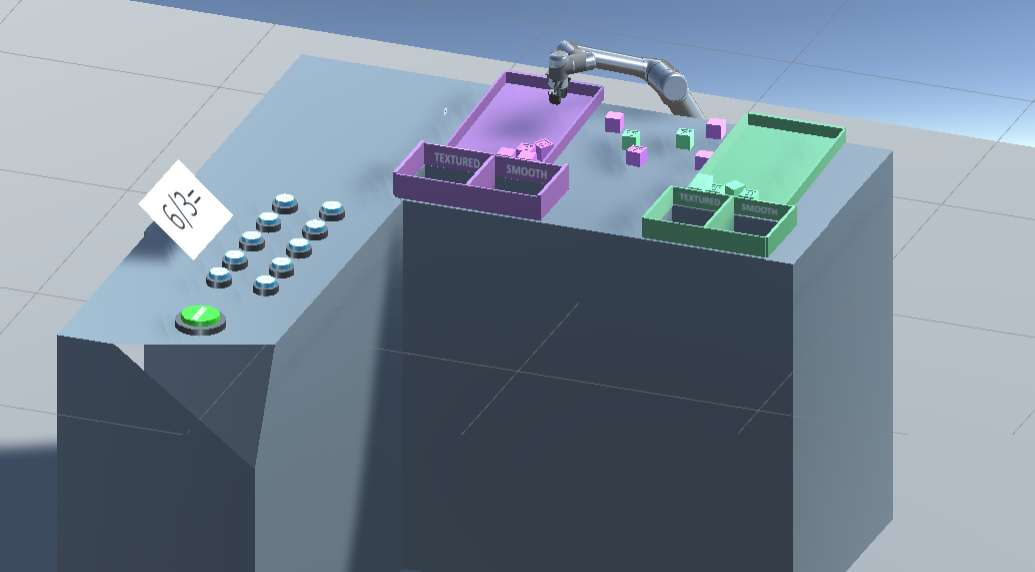}
    \caption{Robot Dual-task. The user solves math problems using the buttons on the platform (left) while supervising a cobot that sorts cubes into different trays by color (right).}
    \label{fig:dualTask}
\end{figure}

\textbf{Collaboration Behind a Wall}
We developed an environment in Unity for a study done in \ac{vr}involving collaboration with a cobot that was sitting behind a visual barrier (a "wall", as seen in fig. \ref{fig:behindthewall}) and the participants had the task of responding to the robot's actions by interacting with objects in the environment. The main goal of the work was to assess how helpful different feedback modalities were in terms of easing the task's cognitive load and enhancing user experience. The team member building the environment was a bachelor student with no prior Unity3D experience at the beginning of the work. After learning how to use Unity3D and connect it to ROS, the team member managed to get a working understanding of Unity and ROS integration within three weeks. Because the study involved a hidden cobot, the tutorial for the study began by showing the participants a cobot sitting behind a transparent glass-like wall. While collaborating with the cobot in the tutorial mode, the wall became more opaque until finally the robot was completely occluded. The study involved audio,visual, and haptic feedback related to the collaborative task, as well as a combination of all three. The study relied on a simulated Niryo One~\footnote{https://niryo.com/fr/product/niryo-one/} robot. Fig. \ref{fig:behindthewall} shows a screenshot of the Unity3D environment.

\textbf{Modular Robots}
For an interdisciplinary study involving computer science and mechanical engineering about modular configurable robots, we built an environment where modules are made of prefabs that can click together. Examples of modules include platform blocks, robotic arms with different configurations, and end-effectors. This gives an overview about the outcome in terms of the robot's shape, dimensions, and functionalities. The 3D models used in simulation of all modules were based on CAD models developed by mechanical engineers working on the project. Simulation of the robot module prefabs did not have to wait until the non-standard physical prototypes for the different modules were built. A screenshot of the work-in-progress can be seen in fig. \ref{fig:modularRobot}.

\textbf{Dual-task Supervision}
We developed another environment (fig. \ref{fig:dualTask}) for a study aimed at exploring the use of different interaction modalities to reduce the cognitive load when supervising a cobot as a secondary task. The user had the task of solving a series of math problems using the buttons on the platform (depicted on the left) while supervising a cobot that is separating colored cubes into different trays by color (seen on the right). The cubes were automatically generated and dispensed, and the behaviour of the robot was fully autonomous, and the participants had to interact with various elements of the environment to perform the required task. This study relied on simulating a Universal Robots e-series robot~\footnote{https://www.universal-robots.com/de/e-series/}.
\section{Discussion}~\label{sec:discussion}
We have demonstrated the use of simulating a robot by these case studies. Based on our conducted studies, we derive the following advantages for simulating a robot rather than buying or building one.

First, the ease of development of simulations. Tools like ROS and Unity3D are offered online for free. This cuts down on the cost of building a simulated environment, whereas actually buying and mounting a robot for a study would cost thousands of euros. This would be in addition to the cost, effort, and time of the logistics involved in transporting the robot to the study location. For example, in our dual-task study (fig. \ref{fig:dualTask}) building the machinery needed for dispensing the cubes, as well as the wiring for the screens and math problems would have been more costly, complicated, and tedious than building it in simulation. Moreover, a simulated robot does not require the space a real robot would need to operate freely without colliding with its environment. The lack of limitation on space makes it easier for a simulated robot to be integrated in virtual environments and studies.

Second, the lower barrier on conducting studies. The cost of mixed reality headsets is lower compared to the cost of a functional cobot, which lowers the financial barrier. A study done with a simulated robot in a virtual setting can be done remotely, provided the user has access to the proper hardware. This lowers the barrier on mobility and removes location restriction. This is useful in situations where a global event restricts mobility, e.g. the COVID-19 pandemic.

Third, the ability to test novel designs. Our finding from the study involving the simulated modular robot was that we did not have to wait until the modules are physically built to begin experimenting. Because the design is not according to a set standard, the mechanical engineers involved in the project had to build the first prototype from scratch, which involved a process of curating the components, measuring, cutting, and assembling the different modules in a functional way. It took less time to build the modules' virtual counterparts and simulate their behavior.

Fourth, the flexibility of the environment. We had more control on the environment in our virtual studies than we would have in a real setting. Working space layout and dimensions were not a constraint for building different settings in terms of layout and size for the studies.

Fifth, repeatability and modularity. A simulated robot would be easier to replicate for a different study, as opposed to physically moving the robot or buying a replica. This allows easier access to validation studies, as well as studies conducted in multiple locations simultaneously. This also allows a modular approach to studies were components from one study could easily be merged or adapted with another.

Studies have shown that a virtual setting can yield results that are as valid as those yielded by an in-situ study.~\cite{Voit2019OnlineArtifacts}, and that \ac{vr} could be beneficial especially for early prototypes~\cite{Mobach2008DoWorlds}. The findings by Matsas et al.~\cite{Matsas2018PrototypingReality} included the benefit of using VR for greater control over the environment and more rapidly changing the conditions. Study results also show that simulated robots in \ac{vr} are suitable for experiments using audiovisual modalities for interaction between humans and robots~\cite{krenn2021s}.

\section{Future Direction and Research Questions}
There are some limitations of simulated robots. For example, a real robot offers a way to test the actual and real value of a lot of the experiments and research when it is carried out in its intended setting. Moreover, using a real robot is more conducive than a simulation for work towards certification and quality assurance. With a real robot, the user has a better sense of the actual real interaction, and can better evaluate aspects relating to sense of safety and trust. When the researcher is faced with the choice between a study in simulation vs. a real robot, it is crucial to use the proper tool evaluate the metrics of interest. It could be possible that predictability, transparency can be equally important in both settings. However, a simulated setting may have additional and specific emphasis on immersion and perceived realism, whereas A real setting could have more emphasis on trust, especially with apprehensive users. The limitations of both settings are grounds for future studies. Based on our findings from the studies we developed, as well as the points discussed in the previous sections, we believe that the focus of future research should be on answering the following research questions:
\textbf{What conditions make better use of simulated robots?} More research should go into guidelines for designing user studies that help researchers decide whether to use simulation, depending on the study conditions and metrics.
\textbf{What are the limitations of both simulation and reality in HRI?} To address the limitations of both aspects, further research should go into how the different methods can complement each other. This should also involve validation studies done in both settings to help derive guidelines for the design of future experiments.

\balance

\balance
\bibliographystyle{ACM-Reference-Format}
\bibliography{references}

\end{document}